\documentclass[11pt,a4paper]{article}
\usepackage[utf8]{inputenc}
\usepackage[T1]{fontenc}
\usepackage[hyperref]{emnlp2018}
\usepackage{times}
\usepackage{latexsym}
\usepackage{graphicx}

\usepackage{url}
\usepackage{booktabs}

\usepackage{algorithm}
\usepackage[noend]{algpseudocode}
\usepackage{amsmath}
\usepackage{amsfonts}

\aclfinalcopy 

\newcommand{\chris}[1]{}

\newcommand{\seb}[1]{}

\newcommand{\kargmax}{\operatornamewithlimits{k-argmax}}

\title{Off-the-Shelf Unsupervised NMT}

\author{Chris Hokamp\textsuperscript{1} \and Sebastian Ruder\textsuperscript{1,2} \and John Glover\textsuperscript{1} \\
  \textsuperscript{1}Aylien Ltd., Dublin, Ireland \\
  \textsuperscript{2}Insight Centre, NUI Galway \\
  \texttt{\string{chris,sebastian,john\string}@aylien.com}\\}

\date{}

\begin{document}
\maketitle

\begin{abstract}

We frame unsupervised machine translation (MT) in the context of multi-task learning (MTL), combining insights from both directions and draw connections to policy-based reinforcement learning.
We leverage off-the-shelf neural MT architectures to train unsupervised MT models with no parallel data and show that such models can achieve reasonably good performance, competitive with models purpose-built for unsupervised MT. 
Finally, we propose improvements that allow us to apply our models to English-Turkish, a truly low-resource language pair. 

\end{abstract}

\section{Introduction}
\label{sec:introduction}

Recent innovations in neural sequence-to-sequence (s2s) models \cite{vaswani-transformer:2017,wu2016google,Miceli-Barone-DeepNMT:2016}
have improved the state of the art in Machine Translation (MT). 
However, the major obstacle to better MT performance for most language pairs is still the lack of high-quality parallel datasets. Given that there is often no shortage of monolingual data, \textit{unsupervised} machine translation has recently become an attractive area of research \cite{Artetxe2018a,Lample2018}.
%
Current unsupervised MT architectures require purpose-built loss functions and model architectures that are specifically designed for the unsupervised setting. In this work we take a step back and experiment with a different approach; training a single model, or a pair of models, with no additional parameters or loss functions beyond the vanilla sequence-to-sequence model.
Our approach is attractive because it allows us to seamlessly leverage any off-the-shelf MT system and we show results competitive with the state-of-the-art for unsupervised MT. Current unsupervised MT methods rely on bilingual dictionaries for initialization, and we also investigate settings for bilingual dictionary induction which make our models more robust. Finally, we report results for English$\leftrightarrow$Turkish, a truly low-resource language pair.





\section{Multi-Task Unsupervised MT}
\label{sec:multi-task-mt}

Multi-task learning~\cite{Caruana1993} encourages the creation of representations that are shared among related tasks, potentially leading to better model generalization. Intuitively, knowledge can be transferred between similar tasks, resulting in better performance and reduced data sparsity over models trained for each task individually.
Similar to work on learning multi-way translation models \cite{firat-multi-way:2017,googleNMT}, recent work on unsupervised MT can be viewed through the lens of multi-task learning.
In particular, existing approaches make use of two types of tasks: 

\begin{itemize}
    \item[] $\mathcal{T}_1: \mathbf{X \longrightarrow Y} $ (Translation)
    \item[] $\mathcal{T}_2: \mathbf{\hat{Y} \longrightarrow Y} $ (Autoencoding)
\end{itemize}
Here we use $\mathbf{X}$ and $\mathbf{Y}$ to denote two distinct domains, in our case text from two different languages, and $\mathbf{\hat{Y}}$ is a corrupted version of $\mathbf{Y}$.

\paragraph{Translation}
In the unsupervised setting, where we only have access to monolingual data, the obstacle in practice is how to obtain the source translations $ \mathbf{X} $ of the target data $ \mathbf{Y} $. Recent work on unsupervised bilingual dictionary induction \cite{Conneau2018,Artetxe2018a} has shown that a reasonably good word- or phrase-level dictionary can be used to obtain a noisy version of $\mathbf{X}$, denoted $\mathbf{\hat{X}}$, by mapping each word or phrase in $\mathbf{Y}$ to its corresponding word or phrase in $\mathbf{X}$, according to the dictionary. $\mathbf{\hat{X}}$ may initially be a poor approximation of $\mathbf{X}$, because it does not have the expected ordering of the source task, and because the dictionary is noisy, and does not take context into account when performing the mapping. However, $\mathbf{\hat{X}}$ can be iteratively refined by jointly learning the inverse of $\mathcal{T}_1$ and $\mathcal{T}_2$, and updating the translation model with the newly generated instances. In practice, we thus also perform the same tasks in the other direction, $\mathcal{T}_3: \mathbf{\hat{Y} \longrightarrow X} $ and $\mathcal{T}_4: \mathbf{\hat{X} \longrightarrow X} $.

\paragraph{Autoencoding}
Several studies have recently shown that including monolingual data in Neural MT (NMT) training curricula can improve performance \cite{Gulcehre:2015:monolingual,Sennrich-Backtranslation:2015,curry:monolingual:2017}. Monolingual data has long been a key part of Statistical MT (SMT) systems \cite{koehn2010smt} 
and can be included into multi-task systems using the \textit{autoencode} task. To make the task more challenging and avoid over-fitting, noise may be added to the input of the auto-encoding task by scrambling or dropping tokens \cite{Lample2018}.



Therefore, training an unsupervised model from two disjoint sets of monolingual data entails the following high-level steps: 1) obtain bilingual word or phrase dictionaries in both translation directions (\textsection \ref{sec:bdi-and-smt}); 2) generate synthetic data $\mathbf{(\hat{X}, Y)} $ and $ \mathbf{(\hat{Y}, X)}$ (\textsection \ref{sec:synthetic}); 3) initialize model(s) using synthetic data for the translation tasks, along with monolingual data for the auto-encoding tasks (\textsection \ref{sec:init}); 4) train initialized models using back-translation (\textsection \ref{sec:training}). We describe each of these in the following sections.

\subsection{Unsupervised Bilingual Dictionary Induction}
\label{sec:bdi-and-smt}

In order to boostrap $\mathbf{\hat{X} \longrightarrow Y} $ and $ \mathbf{\hat{Y} \longrightarrow X} $, we learn an unsupervised bilingual lexicon of word-to-word translations. Similar to \citet{Sogaard2018}, we found the method by \citet{Lample2018} to only work for closely related languages.

For a truly low-resource language pair 
we found it essential to use a seed dictionary of identical strings shared by both vocabularies \cite{Sogaard2018} instead of adversarial training \cite{Conneau2018}. In addition, when extracting the bilingual dictionary, we restrict extracted pairs to those within a certain frequency rank window of one another. We also use n-grams in the bilingual dictionary \cite{Lample2018-smt}.





\subsection{Synthetic Data Generation} \label{sec:synthetic}

Two methods are used to generate synthetic training example pairs for jump-starting training: (1) \textbf{word-translation}, which simply maps each token in $ \mathbf{X} $ to $ \mathbf{Y} $, if it exists in the bilingual dictionary, and vice versa \cite{Lample2018}; and (2) \textbf{smt-translation}, which uses a phrase table built from the bilingual lexicon, along with a monolingual language model, to generate synthetic translations using an SMT decoder \cite{Lample2018-smt}. In order to obtain the translation probability $ p(e|f) $ for the phrase table, we select the top-$k$ words $Y_i$ for source word $ \mathbf{x}_{i} $ based on cross-domain similarity local scaling (CSLS) \cite{Conneau2018} as follows:

\begin{equation}
Y_{i} = \kargmax\limits_{\mathbf{y}_{j} \in \mathbf{Y}} \mathrm{CSLS}(\mathbf{x}_{i}, \mathbf{y}_{j}).
\label{eq:top-k-csls}
\end{equation}
\noindent We then obtain the normalized translation probability for each pair $ (\mathbf{x}_{i}, \mathbf{y}_{j})$ with Eq. \ref{eq:csls-softmax}:
\begin{equation}
p(\mathbf{y}_{j} | \mathbf{x}_{i}) = \frac{e^{\beta * \mathrm{CSLS}(\mathbf{x}_{i}, \mathbf{y}_{j})}}{\sum\limits_{\mathbf{y}_{k} \in Y_i} e^{\beta * \mathrm{CSLS}(\mathbf{x}_{i}, \mathbf{y}_{k})}},
\label{eq:csls-softmax}
\end{equation}
\noindent where $\beta$ is a temperature hyperparameter that we fix at $ \beta = 10 $. The other component of the SMT model is a 5-gram language model trained on the target language in each direction. We use word-translation or smt-translation to translate each monolingual dataset into the other language.\footnote{Note that the actual monolingual data is always the \textbf{target} side of each synthetic training example. In other words, we create training data $ (\mathbf{\hat{X}, Y}) $ and  $ (\mathbf{\hat{Y}, X}) $, but not $ (\mathbf{Y, \hat{X}})$ or $ (\mathbf{X, \hat{Y}})$.}

\subsection{Initialization} \label{sec:init}

We now train initial NMT model(s) on the synthetic translation task data, combined with the autoencode task data, which is created by scrambling segments from each language\footnote{We use the same approach as \citet{Lample2018}, dropping out words with probability $= 0.1$, and shuffling word order within a window of size $= 3$ from each word's original index.}.

\subsection{Training} \label{sec:training}

Finally, we use the initial models to jump-start unsupervised training with dynamic epochs, which generate translation task data by back-translating. During unsupervised training, short training epochs are generated by asking the current model to backtranslate data from $\mathbf{X \longrightarrow \hat{Y}} $, $\mathbf{Y \longrightarrow \hat{X}} $, training with this data, then updating the current model. As using the unsupervised validation metric of \citet{Lample2018} is time-consuming, we validate the number of epochs on the first run and fix the number of epochs for all other runs. The pseudocode for our unsupervised MT method is given in Algorithm~\ref{alg:unsupervised_mt}.


\paragraph{Connection to policy-based reinforcement learning}

%
%
\begin{algorithm}[t]
\caption{Our approach to unsupervised MT.}
\begin{algorithmic}
\small
\State{Initialize epoch counter $E\gets 0$}
\State{Initialize machine translation model weights $\theta$}
\Repeat{}
    \State{Build a new dataset (sample tasks, run translation model with weights $\theta$)}
    \State{Initialize batch counter $K\gets 0$}
    \State{Initialize MT weights $\hat{\theta}\gets \theta$}
    \Repeat{}
        \State{Compute grad $\nabla_{\hat{\theta}} \approx \frac{1}{N} \sum_{i=1}^{N} \nabla_{\hat{\theta}} \log P_{\hat{\theta}}(y_i | x_i)$}
        \State{Update parameters $\hat{\theta}\gets \hat{\theta} + \alpha \nabla{\hat{\theta}}$}
    \Until{$K > K_{\max}$}
\State{Update machine translation model weights $\theta\gets \hat{\theta}$}
\Until{$E > E_{\max}$}
\end{algorithmic}\label{alg:unsupervised_mt}
\end{algorithm}

Our approach can also be viewed as a form of policy-based reinforcement learning (RL)~\cite{Williams1992,Sutton1999}.
From this viewpoint, our translation model is an instance of a policy $\pi_\theta$, that takes actions $a$ (predicting the next word in the target sequence $y$) based on observations $s$ (the source sequence $x$, and any translations produced up to the current timestep).
We sample from the policy $\pi_\theta$ 
and then perform a gradient update step. In particular, analogous to RL, the data encountered by our model is non-stationary, which was also observed in \citet{Xia2016}.

%
%

\subsection{Models}

We test our method with two types of models: s2s with attention \cite{bahdanau2014neural} and the transformer model \cite{vaswani-transformer:2017}. We introduce two ways of re-purposing these models for unsupervised NMT: 

\paragraph{Synchronized} Two models are trained in parallel, one for $ \mathbf{X \longrightarrow Y} $ and another for $ \mathbf{Y \longrightarrow X} $, each supporting the auto-encoding and translation task for one target language. Each model generates new training instances for the other one, for example, $ \mathbf{X \longrightarrow Y} $ takes monolingual samples from $ \mathbf{X} $, and produces outputs $ \mathbf{\hat{Y}} $. The resulting $ \mathbf{(\hat{Y}, X)}$ outputs are then used as translation task training examples for the  $ \mathbf{Y \longrightarrow X} $ system. 

\paragraph{Multi-task} A single model is trained to do all four tasks. Input data from each of the four tasks is pre-pended with a special token indicating the target task that should be used to produce the output. In this setting, the model produces its own training data intrinsically, using the same process described for the synchronized setting.

Note that, in the synchronized setting, we do not need to explicitly indicate tasks to the model, since each model only decodes into one language. In the multi-task setting, we prefix inputs with a special token indicating the target language that should be decoded. 

\section{Experiments}
\label{sec:experiments}

We conduct two sets of experiments to test the viability of learning an NMT system in an unsupervised setting, using monolingual, non-parallel data only, and off-the-shelf s2s models, with \textbf{no modification} to model source code. The first set of experiments follows \citet{Lample2018}, studying the $ EN \leftrightarrow FR $ language pair, using the Multi30k dataset \cite{bojar-EtAl:2017:WMT1}. The second set of experiments is an attempt to establish a realistic unsupervised baseline for $EN \leftrightarrow TR$, a truly low-resource language pair with little typological similarity. 


\begin{table}[ht]
\centering
\begin{tabular}{p{0.5\linewidth}p{0.2\linewidth}p{0.2\linewidth}}
\toprule
 \textsc{System} & \textsc{FR-EN}  \\ 
 \midrule
 \multicolumn{2}{c}{\textbf{\textsc{Word Translation}}} \\
 word-for-word & 13.48 \\
 \multicolumn{2}{c}{\textbf{\textsc{NMT Initialization}}} \\
 Baseline s2s & 20.85  \\
 +autoencode  & \textbf{22.57}  \\
 +scramble    & 21.12  \\
 +pretrained-embeddings (trainable) & 22.55  \\
 +pretrained-embeddings (static) & 21.8  \\
\bottomrule
\end{tabular}
\caption{BLEU scores for \textsc{fr-en} on the Multi-30k development set, training with the word-translation and (optionally) autoencode tasks, translating in one direction.}
\label{tab:fr-en-ablation}
\end{table}

\subsection{EN-FR}

\paragraph{Initialization}

In order to determine reasonable settings, we trained a single \textsc{FR-EN} model for the \texttt{multi30k} dataset, using the baseline s2s with attention model in Marian \cite{junczys2018marian} and $ \mathbf{\hat{FR} \longrightarrow EN} $ data created using the word-translation method. We present results in Table \ref{tab:fr-en-ablation}. Using trainable pretrained embeddings achieved slightly lower results than the best configuration, but converged much faster. We thus choose this configuration for all future experiments.

\begin{table}[ht]
\centering
\begin{tabular}{p{0.53\linewidth}p{0.17\linewidth}p{0.17\linewidth}}
\toprule
 \textsc{System} & \textsc{FR-EN} & \textsc{EN-FR} \\ 
 \midrule
 \multicolumn{3}{c}{\textbf{\textsc{Word Translation}}} \\
 word-for-word & 13.48 & 8.18  \\
 \multicolumn{3}{c}{\textbf{\textsc{NMT Initialization}}} \\
 Initialization (multi-task) & 18.56  & 21.46  \\
 Initialization (sync) & 26.6 & 23.4 \\
 \multicolumn{3}{c}{\textbf{\textsc{NMT Unsupervised}}} \\
 Unsupervised (multi-task) & 18.71  & 18.54  \\
 Unsupervised (sync) & \textbf{31.15} & \textbf{30.2} \\
\bottomrule
\end{tabular}
\caption{Initialization and Unsupervised training results for \texttt{multi30k} development set.}
\label{tab:en-fr-synchronized-results}
\end{table}

\begin{table}[ht]
\centering
\begin{tabular}{p{0.53\linewidth}p{0.17\linewidth}p{0.17\linewidth}}
\toprule
 \textsc{System} & \textsc{FR-EN} & \textsc{EN-FR} \\ 
 Ours & 28.45 & 27.57 \\
 Lample et al. 2018 & \textbf{32.07}  & \textbf{32.76} \\
\bottomrule
\end{tabular}
\caption{Comparing our best model with \citet{Lample2018} on the \texttt{multi30k} test set. We note that they do not specify their tokenizer or BLEU score computation method, whereas we use detokenized, lowercased BLEU, as computed by \texttt{multi-bleu-detok.perl}.}
\label{tab:en-fr-lample-comparison}
\end{table}

\paragraph{Unsupervised Training}

For $ EN \leftrightarrow FR $, we follow \citet{Lample2018} in preparing the datasets. See appendix \ref{sec:appendix:en-fr-training} for configuration details.

The two component models in the synchronized setting use the default s2s with attention, whereas the multi-task model uses the Transformer architecture \cite{vaswani-transformer:2017}, as implemented in Marian \cite{junczys2018marian}\footnote{Note that source and target embeddings must be shared in the multi-task setting}. 

Results are presented in table \ref{tab:en-fr-synchronized-results}. Surprisingly, we observe that the synchronized s2s models perform much better than the multi-task transformer, especially during the unsupervised training phase. We hypothesize that the Transformer model quickly overfits to the autoencoding task our this small dataset. Table \ref{tab:en-fr-lample-comparison} compares the best system with \citet{Lample2018}. Their model, which includes a special adversarial loss and is designed specifically for NMT, is a few BLEU points better, but our simple synchronized models still perform remarkably well.

\subsection{Case study: Unsupervised MT on a truly low-resource language pair} \label{sec:low-resource}

In order to experiment in a more challenging setting, we choose English-Turkish (EN-TR). We use only 2.5 million randomly selected segments from the \texttt{news-crawl} portion of the monolingual training data provided by the WMT task organizers\footnote{\url{http://www.statmt.org/wmt18/translation-task.html}}. 

\subsubsection{Initialization}

For \textsc{EN-TR}, we experimented with SMT initialization, hypothesizing that the language model component of the SMT decoder might be able to correct issues due to the drastic differences in word order between the languages (see appendix for examples of system outputs).

\begin{table}[ht]
\centering
\begin{tabular}{p{0.6\linewidth}p{0.2\linewidth}}
\toprule
 & \textsc{en-tr} \\ \midrule
Unsupervised* & 32.71 \\
Supervised* & 39.22 \\\hline
Fasttext vectors, supervised (default) training & 45.80 \\
Max Rank 20000 & 48.80 \\
Max Rank 30000 & \textbf{48.93} \\
\bottomrule
\end{tabular}
\caption{Bilingual dictionary induction results. *: Results are from \cite{Sogaard2018} and not directly comparable. "Max Rank" indicates the maximum frequency rank for word pairs in Procrustes refinement.}
\label{tab:bdi-results}
\end{table}

The first challenge for this language pair was to find a way to do unsupervised alignment of the embeddings in a way that resulted in a reasonably good dictionary. Table \ref{tab:bdi-results} shows the results of several experiments. By increasing the maximum rank of the extracted pairs, and using the supervised training paradigm, initialized with identically spelled strings \cite{Sogaard2018}, we obtain a significant performance boost over the baseline.

\begin{table}[ht]
\centering
\begin{tabular}{p{0.50\linewidth}p{0.15\linewidth}p{0.15\linewidth}}
\toprule
 \textsc{System} & \textsc{en-tr} & \textsc{tr-en} \\ 
 \midrule
 \multicolumn{3}{c}{\textbf{\textsc{Bilingual Dictionary}}} \\
 word-for-word & 1.49 & 2.29 \\
 SMT & 0.92 & 1.94 \\
 \multicolumn{3}{c}{\textbf{\textsc{NMT Initialization}}} \\
 Baseline s2s + filtered & 1.52 & 2.75 \\
 Baseline s2s + unfiltered & 1.55 & 3.12 \\
 Transformer + unfiltered & 2.05 & 3.13 \\
 \multicolumn{3}{c}{\textbf{\textsc{NMT Unsupervised}}} \\
 Transformer & \textbf{4.42} & \textbf{6.26} \\
\bottomrule
\end{tabular}
\caption{\texttt{newsdev2016} lowercased, detokenized BLEU scores for \textsc{EN}$\leftrightarrow$\textsc{TR}, computed with \texttt{multi-bleu-detok.perl}. "filtered" indicates limiting training data to those pairs which get at least 40 BLEU when round-tripped through both SMT systems.}
\label{tab:en-tr-dev-bleu-scores}
\end{table}

Configuration details are given in appendix \ref{sec:appendix:en-tr-training}. Table \ref{tab:en-tr-dev-bleu-scores} shows the results for word translation and SMT, initialization, and unsupervised training. To the best of our knowledge, these are the first reported results on completely unsupervised MT for \textsc{EN-TR}. On the \texttt{newstest2016} test set, our system also achieves BLEU scores of 5.66 for $\mathbf{EN \rightarrow TR}$ and 7.29 for $ \mathbf{TR \rightarrow EN}$. Given that \citet{curry:monolingual:2017} report a BLEU score of 9.4 for their $\mathbf{EN \rightarrow TR}$ baseline, which uses all available parallel training data, we consider these results quite encouraging.

\section{Conclusions}
\label{sec:conclusions}

We have framed unsupervised MT as multi-task learning, and shown that carefully designed experimental setups can achieve good performance on unsupervised MT with no changes to existing s2s models. Thus, any existing NMT framework could already be used to train unsupervised MT models. We compared to existing purpose built unsupervised MT systems, and established a strong baseline for \textsc{EN-TR}, a truly low-resource language pair. In future work we hope to focus upon including more than two languages in a single model, as well as experimenting with ways to improve initialization. 


\bibliography{acl2018}
\bibliographystyle{acl_natbib_nourl}

\clearpage
\appendix

\section{\textsc{EN-FR} Training Configuration}
\label{sec:appendix:en-fr-training}

We select alternating segments from the English and French portions of the Multi30k corpus, so there are no parallel segments in the two resulting monolingual corpora. 

We test models in both the \textbf{synchronized} and \textbf{multi-task} settings described above. During unsupervised training, we replace the data-generating models with a new model after 30000 training examples have been seen, and train until we have seen 500000 examples total.

To create the bilingual dictionaries for word translation, we use the pretrained fasttext embeddings\footnote{https://fasttext.cc/}, which are then aligned using MUSE. The resulting aligned embeddings achieve $ 81.86 $ $ \mathbf{P@1} $ on the evaluation dataset used by \citet{Conneau2018}.

For the multi-task setting, because we are training a single model which can translate in both directions, we also use 2.5 million sentences from the News Crawl 2017 monolingual EN and FR datasets\footnote{\url{http://www.statmt.org/wmt18/translation-task.html}} to create a \textbf{shared} vocabulary using \textbf{wordpiece} encoding with 32000 symbols \cite{googleNMT}, and train a \textbf{shared} embedding using Fasttext. 

All models are trained for 10 epochs over the training data, batches are evenly sampled from the auto-encoding and translation tasks.

\section{\textsc{EN-TR} Training Configuration}
\label{sec:appendix:en-tr-training}

Using the best unsupervised alignment method from table \ref{tab:bdi-results}, we extract an SMT phrase table by mapping each word to its $K$ most similar words in the other language, using the technique discussed in section \ref{sec:synthetic}. When extracting the dictionary, we require that extracted $\mathbf{SRC \rightarrow TRG}$ mappings are within frequency rank $\pm10000$ of one another. For each language, we also include the top 100000 n-grams up to length 5 in the phrase table, following the technique of \citet{Lample2018-smt}. We train a 5-gram language model for each language using KenLM \cite{Heafield:2011:kenlm}, and produce the $ (\mathbf{\hat{EN}, TR}) $ and $ (\mathbf{\hat{TR}, EN}) $ by decoding in both directions with Moses \cite{Koehn:2007:moses}.

We use a shared wordpiece vocabulary of 8000 symbols, because we hypothesize that the morphological complexity of Turkish will be better captured by a token set which aggressively segments words. 

The initialization models are trained until all SMT-generated data has been seen once, evenly sampling from the translation and autoencoding tasks. The unsupervised model is trained until it has seen 5,000,000 examples, updating the data-generating model every 30,000 examples. 

Evaluation is done on the full WMT 2016 \textsc{EN-TR} development and test sets, using lowercased, detokenized BLEU. This should make our experimental settings easy to replicate, and put a lower bound on what is achievable with more training data, or more sophisticated initialization methods.

\section{Example Outputs}
\label{sec:appendix:translation-examples}

Table \ref{tab:system-output-examples} provides some examples of the output of our models.

\begin{table*}[ht]
\centering
\begin{tabular}{p{0.2\linewidth}p{0.8\linewidth}}
\toprule
Source & bir başka öğrenci mikel sykes ise Lamb'in kendisine 2014-15 öğrenim yılı sonlarında stresle boğuştuğunu söylediğini belirtti. \\
Reference & another student, mikel sykes, said Lamb told him he was dealing with stress at the end of the 2014-15 academic year. \\
word-for-word & a other student darri maccartney and mumford '' him - graduate year in sometime by reducing visibly telling stated. \\
SMT & for david gilmour - year graduate student and another sometime visibly shocked by reducing hesitation mumford said.\\
Transformer (initialization) & another student, mike sykes, acknowledged telling him he had been under pressure while learning him in the past two years. \\
Transformer (unsupervised) & another student, mikel sykes, said lamb's learning was under pressure last year while telling him to drown him in 2014-15. \\
\midrule
Source & zaman zaman çok şiddetli çatışmalara sahne olan ilçede vatandaşlar ilçeyi terk ediyor.\\
Reference & residents are leaving the district, which has occasionally witnessed very violent clashes. \\
word-for-word & time time extremely severe insurgent stage whose titagarh citizens towns, leave should. \\
SMT & citizens should leave time onstage towns , has seen a very intense bloodshed \\
Transformer (initialization) & the violence is a very violent time for citizens to abandon the town 's stage. \\
Transformer (unsupervised) & time to see more violent clashes in the towns where citizens were abandoned. \\
\midrule
Source & son çeyrekte , yer hizmetleri geliri \% arttı ancak işletme geliri büyük paketler ve dahili sigortadaki yüksek maliyetler nedeniyle temelde durgundu.\\
Reference & In the latest quarter, ground revenue rose 29 percent but operating income was basically flat on higher costs for larger packages and self-insurance.\\
word-for-word & last quarter, place services profits \% increased however management profits huge packets and, but now high affordability due radically.\\
SMT & fourth - quarter profits and revenues increased \%, due in larger packets services business and high affordability , but now durgundu sigortadaki radically.\\
Transformer (initialization) & however, revenues increased by \% in the fourth quarter due to higher profits of \$. bn in basic services and management packages.\\
Transformer (unsupervised) & last quarter, revenue increased \%29 per share, however, to packages of large business income and included higher insurance costs due to basic duration.\\

\bottomrule
\end{tabular}
\caption{Example outputs of different systems for $\mathbf{TR \rightarrow EN}$.}
\label{tab:system-output-examples}
\end{table*}






\end{document}